\documentclass{article}

\usepackage{microtype}
\usepackage{graphicx}
\usepackage{subfigure}
\usepackage{booktabs} % for professional tables
\usepackage{multirow}
\usepackage{hyperref}
\usepackage{xcolor}

\definecolor{fillbluelight}{RGB}{184, 187, 249}
\definecolor{fillgreenlight}{RGB}{184, 249, 195}
\definecolor{fillredlight}{RGB}{249, 184, 184}

\usepackage{hyperref}

\usepackage[archival]{icml2023}

\usepackage{amsmath}
\usepackage{amssymb}
\usepackage{mathtools}
\usepackage{amsthm}

\usepackage[capitalize,noabbrev]{cleveref}

\theoremstyle{plain}

\theoremstyle{definition}

\theoremstyle{remark}

\usepackage[textsize=tiny]{todonotes}

\icmltitlerunning{Can strong structural encoding reduce the importance of Message Passing?}

\begin{document}

\twocolumn[
\icmltitle{Can strong structural encoding reduce the importance of Message Passing?}



\icmlsetsymbol{equal}{*}

\begin{icmlauthorlist}
\icmlauthor{Floor Eijkelboom}{uva}
\icmlauthor{Erik Bekkers}{uva}
\icmlauthor{Michael Bronstein}{ox}
\icmlauthor{Francesco Di Giovanni}{cam}

\end{icmlauthorlist}

\icmlaffiliation{uva}{University of Amsterdam}
\icmlaffiliation{ox}{University of Oxford}
\icmlaffiliation{cam}{University of Cambridge}

\icmlcorrespondingauthor{Floor Eijkelboom}{eijkelboomfloor@gmail.com}

\icmlkeywords{Graph, topology, structure, message passing}

\vskip 0.3in
]



\printAffiliationsAndNotice{}  % leave blank if no need to mention equal contribution

\begin{abstract}
The most prevalent class of neural networks operating on graphs are message passing neural networks (MPNNs), in which the representation of a node is updated iteratively by aggregating information in the 1-hop neighborhood. Since this paradigm for computing node embeddings may prevent the model from learning coarse topological structures, the initial features are often augmented with structural information of the graph, typically in the form of Laplacian eigenvectors or Random Walk transition probabilities. In this work, we explore the contribution of message passing when strong structural encodings are provided. We introduce a novel way of modeling the interaction between feature and structural information based on their tensor product rather than the standard concatenation. The choice of interaction is compared in common scenarios and in settings where the capacity of the message-passing layer is severely reduced and ultimately the message-passing phase is removed altogether. Our results indicate that using tensor-based encodings is always at least on par with the concatenation-based encoding and that it makes the model much more robust when the message passing layers are removed, on some tasks incurring almost no drop in performance. This suggests that the importance of message passing is limited when the model can construct strong structural encodings.
\end{abstract}

\section{Introduction}

Graph Neural Networks (GNNs) \citep{gori2005new, scarselli2008graph} is now one of the most studied and adopted framework for machine learning on graphs. GNNs typically operate on attributed graphs, and learn functions that model complex interactions among the underlying topology and the features \citep{sperduti1994encoding, goller1996learning, bruna2013spectral, defferrard2016convolutional}. Since graphs are a flexible way of representing many data, GNNs are being developed in many scientific and industrial fields alike \citep{stokes2020deep, mirhoseini2021graph, dezoort2023graph,bapst2020unveiling}. Accordingly, understanding what makes a GNN work, when it is needed, and how to best leverage the topological information from the graph in the GNN pipeline are all essential problems of broad interest.

One of the most common classes of GNNs is message passing neural networks (MPNNs) \cite{gilmer2017neural}, where the features associated with each node are iteratively updated based on the $1$-hop neighborhood information in the graph. While this approach directly leverages the sparsity of the underlying graph by only exchanging messages among adjacent nodes, a consequence of this learning procedure is that nodes with similar local structures will obtain similar hidden representations, even when they actually play different roles from a coarser topological scale. These ambiguities, ultimately, impact the expressive power of MPNNs. In fact,  \citet{xu2018how, morris2019weisfeiler} showed that MPNNs are, at most, as powerful as the 1-WL color refinement algorithm \citep{weisfeiler1968reduction} in distinguishing pairs of graphs without features. In tasks where the graph structure is fundamental - e.g. in molecular tasks - this limits the capabilities of the network severely.

A solution to the problem of limited expressive power of MPNNs that minimizes the impact on the efficiency by preserving the message-passing template, amounts to augmenting the initial node features with structural information \citep{bouritsas2022improving, dwivedi2021graph}, 
 and is at the heart of applications of transformers to graphs \citep{ying2021transformers, kreuzer2021rethinking}. This graph-aware feature augmentation is typically based on either the eigenvectors of the graph Laplacian or on the transition probabilities of random walks on the graph. Although the choice of the structural information to be provided is being investigated extensively \citep{rampavsek2022recipe}, {\em how} we combine this information with the input features is a much less studied subject. In fact, the canonical choice in this regard is simply concatenating the node-features and the structural (positional) ones before feeding them into a message-passing layer (often after a suitable encoding). Moreover, while structural information is now known to improve MPNNs, especially on molecular tasks, it is unclear to what extent message-passing is needed, especially if we can extract structural information from the graph in a powerful way.

\paragraph{Contributions.} In this paper, we explore structural encoding in GNNs and the role of the message-passing paradigm when using structural features. First, we introduce a novel way of modeling the interaction between feature and structural information based on their {\em tensor product} -- rather than the conventional concatenation operation -- and compare how these two approaches to structural encoding affect performance in different GNN models. Second, in order to assess the impact of the structural encoding compared to that of the actual message-passing paradigm, we study the effects of reducing the weights in the message-passing layer and number of message passing layers in the model, ultimately entirely removing the message-passing phase.

Our results indicate that using a tensor-product encoding is always on par with the concatenation-based encoding, often slightly outperforming the latter. Moreover, we provide empirical evidence that when excluding the message passing layer from our network, the tensor-based encoding is significantly more robust than its concatenation counterpart. In fact, 
on some tasks, when using a tensor encoding almost no drop in performance is observed, and a GNN consisting of encoder and decoder, {\em without any message passing}, performs on par with GNNs that include message passing even when the number of parameters used is significantly reduced. Furthermore, we observe that when the model is provided with strong structural encodings, there is no major improvement noticeable when using dense message passing layers compared to using very sparsely parametrized ones for both concatenation and tensor encodings. This suggests that {\em the role of message passing layers may often be limited when the model is able to encode structural information in a sufficiently powerful way}.

\section{Preliminaries} 

\paragraph{Graphs and MPNNs.}

Let $\mathcal{G} = (\mathcal{V}, \mathcal{E})$ be an undirected graph with nodes $\mathcal{V}$ and edges $\mathcal{E}$ and let $i \sim j$ denote that nodes $i$ and $j$ are connected, i.e., $(i,j)\in \mathcal{E}$. The adjacency matrix $\mathbf{A}$ encodes the connectivity of the graph since $A_{ij} = 1$ if $i\sim j$ and $A_{ij} = 0$ otherwise, while the degree matrix $\mathbf{D} = \mathrm{diag}(d_1,\ldots,d_n)$, where $d_i = \sum_{j}A_{ij}$, represents the connectivity of each node.

In the deep learning context,  the nodes of the graph are typically endowed with features $\{\textbf{h}_i:\,i\in\mathcal{V}\}\subset \mathbb{R}^{d}$. Graphs lend themselves as useful descriptors for data that live on irregular domains, e.g. molecules or social networks. Graph neural networks (GNNs) are neural architectures that take as input graphs and process the graph structure in a permutation-equivariant way, e.g., to  predict the chemical properties of a molecular graph.

One of the most common classes of GNNs are Message Passing neural networks (MPNNs) \cite{gilmer2017neural}, where the features associated with each node are iteratively updated based on the $1$-hop neighborhood information, i.e.
\begin{equation}\label{eq:mpnn}
\textbf{h}_i^{\ell + 1} = \mathsf{MPNNLayer}(\textbf{h}^{\ell}_i, \{\!\!\{\textbf{h}^{\ell}_j\}\!\!\}_{i \sim j}),
\end{equation}
where $\textbf{h}^{\ell}_i$ denotes the hidden state of node $i$ in layer $\ell$ and $\{\!\!\{\textbf{h}^{\ell}_j\}\!\!\}_{i \sim j}$ is the multiset of adjacent features. Note that the MPNN formalism includes several main instances of GNN architectures such as GCN \cite{kipf2016semi}, GIN \cite{xu2018how}, or SAGEConv \cite{hamilton2017inductive}. % or GAT are there specific types of MPNNs. 
The initial features $\{\textbf{h}_i^{\text{in}}\}$ are typically embedded using some multi-layer Perceptron (MLP), 
$$\textbf{h}_i^{\ell = 0} = \mathsf{Encode}(\textbf{h}_i^{\text{in}}).$$ To get a hidden state representing the entire graph -- which is needed for graph-level tasks -- a permutation-invariant readout map is applied to all final hidden states of the nodes, that is for an MPNN of $L$ layers we have
$$\textbf{h}_{\mathcal{G}} = \underset{v \in \mathcal{V}}{\mathsf{Agg}} ~\textbf{h}^{L}_v,$$
where $\mathsf{Agg}$ is a permutation invariant operator such as sum, mean, or max. 
This state is then passed through a final decoder (often an MLP) to derive  the final prediction computed by the MPNN. 

\paragraph{Augmented MPNNs.}

A now established paradigm to frame the expressive power of GNNs is the graph-isomorphism test. In  fact, \citet{xu2018how,morris2019weisfeiler} proved that MPNNs are, at most, as powerful as the 1-WL test \citep{weisfeiler1968reduction} in distinguishing {\em unattributed} graphs.
%The theoretical limitations of standard MPNNs are commonly understood in the context of the 1-dimensional Weisfeiler-Lehman test (1-WL) on graphs \cite{xu2018powerful}, where the features of the nodes describe the colors and adjacency relations are defined by the edges of the graph (see \citet{kipf2016semi}). 
As a consequence of this analogy, MPNNs typically struggle to recognize (or count) substructures \citep{chen2020can} and will, for instance, assign equal graph-level embeddings to the graphs in \autoref{fig:nonisographs} despite them being non-isomorphic due to the existence of cycles of different length.

\begin{figure}[h]
    \centering
    \begin{tikzpicture}[fill opacity=1, scale=1.5]
		\tikzstyle{point}=[circle,fill=black,inner sep=0pt,minimum size=3pt]
		\node (a)[point] at (0, .25) {};
		\node (b)[point] at (0, .75) {};
		\node (c)[point] at (.45, 0) {};
		\node (d)[point] at (.45, 1) {};
		\node (e)[point] at (.9, .25){};
  		\node (f)[point] at (.9, .75){};

            \node (g)[point] at (1.35, 1) {};
		\node (h)[point] at (1.35, 0) {};
		\node (i)[point] at (1.8, .75) {};
		\node (j)[point] at (1.8, .25) {};

            \node (aa)[point] at (2.4, .25) {};
		\node (ba)[point] at (2.4, .75) {};

            \node (ca)[point] at (2.95, .1) {};
		\node (da)[point] at (2.95, .9) {};
  
            \node (ea)[point] at (3.3, .5) {};
		\node (fa)[point] at (3.75, .5) {};

            \node (ga)[point] at (4.1, .1) {};
		\node (ha)[point] at (4.1, .9) {};
            
            \node (ia)[point] at (4.65, .75) {};
		\node (ja)[point] at (4.65, .25) {};

            \draw[fill=fillbluelight] (a.center) --  (b.center) -- (d.center) -- (f.center) -- (e.center) -- (c.center) -- cycle;
            \draw[fill=fillbluelight] (e.center) --  (f.center) -- (g.center) -- (i.center) -- (j.center) -- (h.center) -- cycle;

            \draw[fill=fillgreenlight] (aa.center) -- (ba.center) -- (da.center) -- (ea.center) -- (ca.center) -- cycle;
            \draw[fill=fillgreenlight] (fa.center) -- (ga.center) -- (ja.center) -- (ia.center) -- (ha.center) -- cycle;

    	\draw[line width=1.15pt] (a.center) -- (b.center);
    	\draw[line width=1.15pt] (a.center) -- (c.center);
    	\draw[line width=1.15pt] (e.center) -- (c.center);
    	\draw[line width=1.15pt] (e.center) -- (f.center);
    	\draw[line width=1.15pt] (d.center) -- (f.center);
    	\draw[line width=1.15pt] (d.center) -- (b.center);
    	\draw[line width=1.15pt] (f.center) -- (g.center);
    	\draw[line width=1.15pt] (e.center) -- (h.center);
    	\draw[line width=1.15pt] (j.center) -- (h.center);
    	\draw[line width=1.15pt] (j.center) -- (i.center);
    	\draw[line width=1.15pt] (g.center) -- (i.center);

    	\draw[line width=1.15pt] (aa.center) -- (ba.center);
    	\draw[line width=1.15pt] (aa.center) -- (ca.center);
    	\draw[line width=1.15pt] (ea.center) -- (ca.center);
    	\draw[line width=1.15pt] (ea.center) -- (fa.center);
    	\draw[line width=1.15pt] (da.center) -- (ea.center);
      	\draw[line width=1.15pt] (ba.center) -- (da.center);
      	\draw[line width=1.15pt] (fa.center) -- (ga.center);
      	\draw[line width=1.15pt] (fa.center) -- (ha.center);
      	\draw[line width=1.15pt] (ia.center) -- (ha.center);
      	\draw[line width=1.15pt] (ia.center) -- (ja.center);
      	\draw[line width=1.15pt] (ga.center) -- (ja.center);

    	% \draw[line width=1.15pt] (da.center) -- (ba.center);
    	% \draw[line width=1.15pt] (fa.center) -- (ga.center);
    	% \draw[line width=1.15pt] (ea.center) -- (ha.center);
    	% \draw[line width=1.15pt] (ja.center) -- (ha.center);
    	% \draw[line width=1.15pt] (ja.center) -- (ia.center);
    	% \draw[line width=1.15pt] (ga.center) -- (ia.center);

            \draw[point] (a) circle (0.04);
		\draw[point] (b) circle (0.04);
		\draw[point] (c) circle (0.04);
		\draw[point] (d) circle (0.04);
		\draw[point] (e) circle (0.04);
		\draw[point] (f) circle (0.04);
		\draw[point] (g) circle (0.04);
		\draw[point] (h) circle (0.04);
		\draw[point] (i) circle (0.04);
		\draw[point] (j) circle (0.04);

            \draw[point] (aa) circle (0.04);
		\draw[point] (ba) circle (0.04);
		\draw[point] (ca) circle (0.04);
		\draw[point] (da) circle (0.04);
		\draw[point] (ea) circle (0.04);
		\draw[point] (fa) circle (0.04);
		\draw[point] (ga) circle (0.04);
		\draw[point] (ha) circle (0.04);
		\draw[point] (ia) circle (0.04);
		\draw[point] (ja) circle (0.04);
            
\end{tikzpicture}
    \caption{Non-isomorphic graphs not distinguished by 1-WL.}
    \label{fig:nonisographs}
\end{figure}
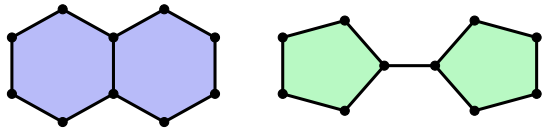

\begin{figure}[h]
	\centering
	
\end{figure}

In tasks such as molecular prediction, identifying substructures and hence breaking ambiguities intrinsic to the 1-WL color refinement, play an important role. To compensate for the shortcomings of classical MPNNs, several approaches have been proposed to design more powerful GNNs \citep{maron2018invariant, keriven2019universal, geerts2022expressiveness}. However, most of these frameworks suffer from often impractical space-time complexity. 

A different approach, amounts to augmenting the standard MPNN formulation in various ways by incorporating  {\bf topological information} from the underlying graph -- in accordance with the conventional nomenclature, we use the terminology `structural features' (or structural encodings). Such augmentation can take many forms, and we outline two broad strategies. One way considers the graphs' structure as explicit features  \cite{bodnar2021weisfeiler}. By considering e.g. rings as explicit features to be learned, we construct more complex neighborhood structures with the consequence of being able to distinguish strictly more graph isomorphisms (see also \citet{morris2019weisfeiler}). Though effective, these methods tend to be computationally expensive and typically require more apriori domain-knowledge to identify beforehand the structures that are likely to be relevant for the task.

A second approach is to include structural information in the node features, meaning that the initial feature and structural information are altered as follows:
$$\textbf{h}_i^{\ell = 0} = \mathsf{Encode}(\textbf{h}_i^{\text{in}}, \textbf{p}_i^{\text{in}}),$$
where $\textbf{p}_i^{\text{in}}$ denotes the structural information of node $i$. Typically, the encodings are formed by first concatenating $\textbf{h}_i^{\text{in}}$ and $\textbf{p}_i^{\text{in}}$ and then passing the resulting state through an MLP. Besides augmenting initial features only, alternative approaches such as updating the structural information through the message passing layers separately, have also been explored \cite{dwivedi2021graph}. Augmenting node features with graph-aware structural information can be particularly useful for graph transformers, where structural information is used as positional information when passing messages \citep{ying2021transformers, kreuzer2021rethinking, rampavsek2022recipe}.

\paragraph{Structural (topological) information.}
Although different strategies for augmenting features with graph-information have been proposed, random-walk transition probabilities and Laplacian-eigenvectors constitute the most adopted classes of structural encodings \citep{dwivedi2021graph, rampavsek2022recipe}. A random walk is a Markov chain supported on the vertices of the graph with transition matrix $\mathbf{R}$ defined by $\mathbf{R} = \mathbf{A}\mathbf{D}^{-1}$ -- if we take any probability distribution over the graph to be a row vector. The matrix $\mathbf{R}$ is typically raised to various powers to find structural information for random walks of different lengths, where either all entries in $\mathbf{R}^k$ are used for pairwise nodes, or just the diagonal entries to only describe landing probabilities of a node to itself (see \citet{li2020distance, dwivedi2021graph} respectively).
Random walks are powerful descriptors of the graph-topology -- since they capture the diffusion properties of the graph -- and are intimately related to the spectrum of the Laplacian \cite{chung1997spectral}. In fact, recently \citet{black2023understanding,di2023over} proved that the phenomenon of over-squashing is likely to occur among nodes that are hard to be visited by a random walk, i.e. with large commute time. 

An alternative to using random-walk features is found in spectral graph theory, i.e. the study of properties of graphs using the eigenvectors and eigenvalues of matrices describing the graph. For example, the eigenvectors of the graph Laplacian are informative about the connectedness of the graph, and as such are commonly used as structural features in GNNs. Recall that the Laplacian $\mathbf{L}$ of a graph is computed by taking the difference between degree and adjacency matrices, i.e. $\mathbf{L} = \mathbf{D} - \mathbf{A}$. For a Laplacian $\mathbf{L}$ with eigendecomposition
$\mathbf{L} = \mathbf{U^\top \Lambda U},$
one can define $\textbf{p}_i^{\text{lap}} := \begin{bmatrix}\mathbf{\tilde{U}}_{i1} & \cdots & \mathbf{\tilde{U}}_{ik} \end{bmatrix},$
where $\mathbf{\tilde{U}}$ is matrix $\mathbf{U}$ reordered such that the $k$th row in $\mathbf{\tilde{U}}$ has the $k$th highest eigenvalue.

\section{Tensor-product structural encoding}

In this section, we introduce an alternative approach to encode structural information in GNNs that leverages tensor products, rather than the typical concatenation operation. We take inspiration from physics, and in particular the description of quantum states, and design an encoding operation that instead of concatenating input features and structural features, takes the tensor product of the representation.

\paragraph{A novel structural encoding.} Formally, the {\em tensor} (or {\em Kronecker}) {\em product} $\otimes$ between matrices $\mathbf{A}$ and $\mathbf{B}$ is the matrix formed by replacing each element in $\mathbf{A}$ with the product of that element with $\mathbf{B}$, i.e.
$$\mathbf{A} \otimes \mathbf{B} = \begin{bmatrix}
    a_{11} \mathbf{B} & \cdots & a_{1n} \mathbf{B} \\
    \vdots & \ddots & \vdots \\
    a_{m1} \mathbf{B} & \cdots & a_{mn} \mathbf{B}
\end{bmatrix}.$$
Therefore, when considering $\textbf{a} \otimes \textbf{b}$ for vectors $\textbf{a} \in \mathbb{R}^d, \textbf{b} \in \mathbb{R}^q$, we find that the product $a_i \cdot b_j$ is the element in index $i \cdot q + j$, i.e. being the vector found by concatenating vectors $\textbf{b}$ scaled for each element in $\textbf{a}$. 

Typically, when constructing an encoding for our input features and structural information, the respective vectors are first concatenated and then projected, i.e.
$$(\textbf{h}_i, \textbf{p}_i) \mapsto \mathbf{W}\begin{bmatrix}
        \textbf{h}_i \\
        \textbf{p}_i
    \end{bmatrix},$$
for some weight matrix $\mathbf{W} \in \mathbb{R}^{d_{\text{hidden}} \times (d + q)}$.
To embed the input features using a tensor product, we simply mimic this approach, i.e.
$$(\textbf{h}_i, \textbf{p}_i) \mapsto \mathbf{W}(\textbf{h}_i \otimes \textbf{p}_i),$$
for some weight matrix $\mathbf{W} \in \mathbb{R}^{d_{\text{hidden}} \times (d \cdot q)}$. The main rationale behind our choice of studying a tensor product representation, is that it is now possible to construct a much higher-dimensional (and hence more expressive) node representation that accounts for both features and graph topology, even by starting from lower-dimensional entities. 

However, notice that the number of inputs to the projection has increased significantly, which is something we account for in the experiments by significantly reducing the input dimension. More generally though, below we explore alternative ways of reducing the number of parameters that are inspired by tensor-product and the phenomenon of entanglement. We note that from now on, in both the discussion and in the experiments, we restrict to the case where the feature and the structural coordinates share the same dimension so that $(\mathbf{h}_i,\mathbf{p}_i)$ is mapped to a vector in $\mathbb{R}^{d_\text{hidden}\times d_\text{hidden}}$. 

\paragraph{Sparsifying message-passing through entanglement.}

It is common for a message-passing layer to update the feature of a state by using a linear projection of the current state combined with the aggregation of the messages from its neighbours, followed by a non-linear activation. Put differently, for many common instances of MPNNs, the $\mathsf{MPNNLayer}$ operation in \eqref{eq:mpnn} can be taken to be an MLP. Accordingly, the vast majority of the parameters used by a GNN enter the MPNN layer through the weight matrices of the MLP. Since the proposed tensor-product encoding might have a larger contribution to the size of the neural network, and we are generally interested in assessing the importance and impact of the message-passing step in the GNN, we introduce a novel way of controlling the expressivity of $\mathsf{MPNNLayer}$ and its parameter count.

Suppose that $\textbf{h}_i \in \mathbb{R}^{d^2}$ for some  $d$ -- as in the cases of $d$-dimensional features augmented with $d$-dimensional structural encoding using a tensor product. We  can then `reshape' $\textbf{h}_i$ to be a $d \times d$ matrix, which we denote by $\mathsf{Mat}(\textbf{h}_i)$. We then propose to approximate the linear layer-projection in $\mathsf{MPNNLayer}$ with a series of small matrices $\{(\mathbf{W}_k, \mathbf{Q}_k)\}_{k=1}^K$ with $\mathbf{W}_k, \mathbf{Q}_k \in \mathbb{R}^{d \times d}$, by left and right multiplying the node-state $\mathsf{Mat}(\textbf{h}_i)$, i.e.
\begin{equation}\label{eq:entanglement_layer}
\textbf{h}_i \mapsto \sum_{k=1}^K \mathbf{W}_k \mathsf{Mat}(\textbf{h}_i) \mathbf{Q}_k.
\end{equation}
The update is then used on the current state and the aggregation over the 1-hop as per \eqref{eq:mpnn} before the action of a pointwise non-linear activation, e.g. for GIN with a one-layer MLP the state $\mathsf{Mat}(\textbf{h}_i)$ would be updated as follows: 

$$\mathsf{ReLU}( \sum_{k=1}^K \mathbf{W}_k ((1 + \epsilon) \cdot \mathsf{Mat}(\textbf{h}_i) + \sum_{j \sim i} \mathsf{Mat}(\textbf{h}_j)) \mathbf{Q}_k).$$

Note that when using the sparse linear layers, the expressivity of the message passing procedure is constrained, and using a larger value of $K$ increases its expressivity. We remark that the proposed sparsification reduces the number of parameters in each MPNN-layer from $d^4$ to $2K \times d^2$ parameters, i.e. a reduction of a factor of  $\tfrac{d^2}{2K}$ parameters in general. 

We note that if the node-state satisfies $\mathbf{h}_i' = \mathbf{h}_i \otimes \mathbf{p}_i$, then reshaping the tensor-product state into a matrix, simply corresponds to writing it as the outer product $\mathbf{h}_i\mathbf{p}_i^\top$, meaning that the sparse layer-update in \eqref{eq:entanglement_layer} takes the form
\begin{align*}
    \textbf{h}_i' \mapsto \sum_{k=1}^K (\mathbf{W}_k\mathbf{h}_i)( \mathbf{Q}_k\mathbf{p}_i)^\top &= \sum_{k=1}^K (\mathbf{W}_k\otimes  \mathbf{Q}_k)(\mathbf{h}_i\otimes \mathbf{p}_i) \\
    &=\sum_{k=1}^K (\mathbf{W}_k\otimes  \mathbf{Q}_k)\mathbf{h}_i'.
\end{align*}
Accordingly, the number $K$ determines the level of `entanglement' between the states $\mathbf{h}_i$ and $\mathbf{p}_i$ at node $i$. In fact, each linear matrix on the tensor (outer) product space can be decomposed as a sum of tensor products of smaller matrices; since any matrix of shape $(d \times d) \times (d \times d)$ can be represented using a linear combination of at most $K = d^4$ of such tensor product matrices, the larger $K$, the more entanglement we can model and hence the more degrees of freedom we can model in the tensor product space. {\em This is a more geometric approach to controlling the size of the weights and hence the number of parameters in the message-passing layer.}

We finally stress that even for states that are not in a tensor-product form, the sparsification introduced in \eqref{eq:entanglement_layer} still makes sense, and will in fact be used for the concatenation case as well in our experiments.

%\textbf{Do we want to mention entanglement here?}

\section{Empirical analysis}

The aim of the experimental evaluation is twofold. First, we evaluate the difference in performance when using tensor encoding versus concatenation encoding by comparing their performances for multiple MPNN layers and datasets. Second, to better understand to what extent the graph structure can be captured without (much) message passing by both types of encodings, a comparison is done in which a low parameter regime for the message passing is considered. The capacity of the message passing is constrained by reducing either the number of weights or the number of layers  -- ultimately removing the message passing altogether. 

To ensure the models considered can embed the structural information well, the input features and structural information are first embedded separately before either concatenation or a tensor product is applied, i.e.
$$\textbf{h}^{\ell = 0}_i = \mathsf{Encode}(\mathbf{W}_h \textbf{h}_i, \mathbf{W}_p \textbf{p}_i),$$
where $\mathsf{Encode}$ is either the concatenation or tensor product encoding and $\mathbf{W}_h, \mathbf{W}_p$ denote linear projections of the input features and structural information respectively. Moreover, the readout is done by summing over the final hidden states and passing the resulting state through a two-layer MLP for the final prediction. 

\begin{table*}[t!]
    \centering
    \caption{Results on Peptides-Struct (MAE) and Peptides-Func (AP). All results are averaged over 3 runs with different seeds.}

        \vspace{2mm}

    \begin{tabular}{rr|cccc}
        \toprule
        \multicolumn{6}{c}{$\mathsf{PeptidesStruct}$ (MAE $\downarrow$)} \\
        \midrule
        &State Type  & full & $K=10$ & $K=1$  & no MP   \\
        \midrule
        \parbox[t]{2mm}{\multirow{3}{*}{\rotatebox[origin=c]{90}{GCN}}}
        &RW-Concat  & 0.212 / 0.255& 0.21 / 0.255 & 0.226 / 0.251  & 0.64 / 0.643 \\
&RW-Tensor & 0.212 / 0.252& 0.206 / 0.254 & 0.223 / 0.249  & 0.249 / 0.268 \\
&Gain  & 1.0 / 1.012 & 1.019 / 1.004& 1.013 / 1.008 & 2.57 / 2.399 \\
        \midrule
        \parbox[t]{2mm}{\multirow{3}{*}{\rotatebox[origin=c]{90}{GIN}}}
        &RW-Concat  & 0.5 / 0.29 & 0.64 / 0.555 & 0.225 / 0.252 & 0.64 / 0.643\\
&RW-Tensor  & 0.239 / 0.257  & 0.498 / 0.309& 0.22 / 0.249& 0.25 / 0.266 \\
&Gain & 2.092 / 1.128  & 1.285 / 1.796& 1.023 / 1.012& 2.56 / 2.417  \\
        \midrule
        \parbox[t]{2mm}{\multirow{3}{*}{\rotatebox[origin=c]{90}{SAGE}}}
        &RW-Concat  & 0.21 / 0.254 & 0.218 / 0.256 & 0.227 / 0.251& 0.64 / 0.643 \\
&RW-Tensor & 0.198 / 0.253& 0.219 / 0.253 & 0.228 / 0.25  & 0.254 / 0.267 \\
&Gain  & 1.061 / 1.004  & 0.995 / 1.012& 0.996 / 1.004&  2.52 / 2.408 \\
        \midrule
        \midrule
        \multicolumn{6}{c}{$\mathsf{PeptidesFunc}$ (AP $\uparrow$)} \\
        \midrule
        &State Type  & full & $K=10$ & $K=1$   & no MP  \\
        \midrule
        \parbox[t]{2mm}{\multirow{3}{*}{\rotatebox[origin=c]{90}{GCN}}}
        &RW-Concat  & 0.906 / 0.613 & 0.877 / 0.598& 0.786 / 0.606  & 0.509 / 0.468\\
&RW-Tensor  & 0.875 / 0.607 & 0.846 / 0.595& 0.784 / 0.593  & 0.766 / 0.586\\
&Gain  & 0.966 / 0.99 & 0.965 / 0.995& 0.997 / 0.979 & 1.505 / 1.252 \\
        \midrule
        \parbox[t]{2mm}{\multirow{3}{*}{\rotatebox[origin=c]{90}{GIN}}}
        &RW-Concat & 0.672 / 0.577 & 0.165 / 0.328& 0.847 / 0.596 & 0.548 / 0.49  \\
&RW-Tensor & 0.921 / 0.611 & 0.923 / 0.601& 0.854 / 0.606  & 0.736 / 0.582 \\
&Gain  & 1.371 / 1.059 & 5.594 / 1.832& 1.008 / 1.017  & 1.343 / 1.188\\
        \midrule
        \parbox[t]{2mm}{\multirow{3}{*}{\rotatebox[origin=c]{90}{SAGE}}}
        &RW-Concat & 0.907 / 0.607 & 0.844 / 0.594& 0.773 / 0.596 & 0.555 / 0.501  \\
&RW-Tensor  & 0.929 / 0.616& 0.842 / 0.607 & 0.776 / 0.601  & 0.724 / 0.579\\
&Gain  & 1.024 / 1.015 & 0.998 / 1.022& 1.004 / 1.008  & 1.305 / 1.156\\
        \bottomrule
    \end{tabular}
    \label{tab:peptides_res}
\end{table*}

To compensate for the number of features in the tensor encodings compared to the case of concatenation, smaller dimensions are provided to the weight matrices $\mathbf{W}_h$ and $\mathbf{W}_p$ when a tensor product is used. Namely, when using a resulting hidden dimension in the layers of $d^2$ features, for tensor the weight matrices embed the inputs to $d$-dimensional vectors whereas in concatenation the matrices embed the inputs to $\frac{d^2}{2}$ features. As such, this comparison is restricted to using hidden dimensionalities of $d^2$ features for some even number $d$ and ensures that both approaches share similar parameters.

The message passing layers considered in this work are GCN, GIN, and SAGEConv. The effect of the structural encodings is compared when 1) not reducing the number of parameters in the linear layers, referred to as `full' message passing, 2) using the sparse linear layers as seen in \eqref{eq:entanglement_layer} for two different levels of entanglement, referred to as `sparse' message passing, and 3) when removing the message passing layers altogether. Note that when doing either sparse or no message passing, we do not compensate for the reduction in parameters in any way, implying that the sparse and no message passing models have significantly fewer weights.
 Moreover, when no message passing is done, the model is reduced to only the encoding and decoding phases, meaning that all the graph structure must be learned through the structural encodings {\em only}. Last, an ablation study is conducted for GCN to compare the effect of reducing both the number of layers and the number of parameters in the models.

For the main experiments, a $4$-layer MPNN architecture with approximately $500$K parameters using full message passing is used as the point of reference for each of the MPNN layer types. The corresponding hidden dimension is then also used when doing sparse message passing and no message passing in which cases the number of parameters is severely reduced. We report the final training and test performance averaged over $3$ runs. We also denote the relative improvement of tensor-encodings (`gain'). In all experiments, we use the structural encoding 
$$\textbf{p}_i := \begin{bmatrix}
    \mathbf{R}_{ii} & \mathbf{R}_{ii}^2 & \cdots \mathbf{R}_{ii}^k
\end{bmatrix},$$
i.e. the diagonal entries of the random walk matrix $\mathbf{R}$ with random walks up to length $20$.

All models are optimized using Adam with a learning rate of $10^{-3}$ using a scheduler halving the learning rate whenever the model does not improve for $25$ epochs. Training is stopped when the learning rate drops below $10^{-5}$. The regression tasks are optimized based on mean average error (MAE) which is the metric reported, whereas the classification is optimized using cross entropy and the reported metric is the average precision (AP). For all main results, we provide standard deviations over the multiple runs in \autoref{app:results_det}, though these are omitted in the main text for visual clarity.

\subsection{Results}

\paragraph{Peptides.}

The Long Range Graph Benchmark is a collection of five graph-based datasets with tasks that are based on long-range dependencies in graphs. Two of these datasets - Peptides-struct and Peptides-func - are multi-class regression and classification in the domain of chemistry. Results are reported as train/test pairs and are provided in \autoref{tab:peptides_res}. These datasets have been introduced as benchmarks for tasks that may exhibit long-range interactions, and have in fact been used by graph-transformers or GNN-models that are designed to reduce over-squashing \citep{rampavsek2022recipe, gutteridge2023drew}.

When considering full weight matrices, a tensor encoding performs at least as well as concatenation, never leading to a significant decrease in test performance. Furthermore, we observe that using sparse linear layers in the MPNNs does not lead to stark decreases in performance in general. For the Peptides-struct task, we observe that both the final train MAE and test MAE are quite similar to the performance when using the approximate layers, suggesting that indeed the {\em role of MPNN layers in this task might not be as crucial -- or at least that over-parameterization of the message-passing layers is redundant} -- something which is emphasized by the fact that even on the train data the model does not learn more when using full weight matrices compared to the approximate layers. On the Peptides-func task, we also notice that the difference between the test performances of the full weights versus the approximate layers is very minimal.

On the Peptides-struct task, we observe that not using MPNN layers at all leads to a stark decrease in performance compared to using the full weights in each layer when the encodings are formed through concatenation, whereas when using a tensor encoding such a decrease in performance is not present. A similar effect is seen in the Peptides-func task, where even though both encoding types perform worse when no MPNN layers are added, this decrease in performance is much stronger when concatenation is used. This suggests that tensor encodings are able to capture more complex information in their encodings, supporting our intuition for using tensor encodings in MPNNs.

\begin{table*}[t!]
    \centering
    \caption{Results on ZINC (MAE). All results are averaged over 3 runs with different seeds.}

    \vspace{2mm}

    \begin{tabular}{rr|cccc}
        \toprule
        \multicolumn{6}{c}{$\mathsf{ZINC}$ (MAE $\downarrow$)} \\
        \midrule
        &State Type   & full  & $K=10$ & $K=1$   & no MP \\
        \midrule
        \parbox[t]{2mm}{\multirow{3}{*}{\rotatebox[origin=c]{90}{GCN}}}
        &RW-Concat  & 0.02 / 0.207& 0.043 / 0.233 & 0.135 / 0.24  & 0.466 / 0.607\\
&RW-Tensor & 0.043 / 0.209& 0.052 / 0.22  & 0.12 / 0.219 & 0.324 / 0.527 \\
&Gain & 0.465 / 0.99  & 0.827 / 1.059& 1.125 / 1.096 & 1.438 / 1.152 \\
        \midrule
        \parbox[t]{2mm}{\multirow{3}{*}{\rotatebox[origin=c]{90}{GIN}}}
        &RW-Concat  & 0.076 / 0.263& 0.04 / 0.281 & 0.116 / 0.279 & 0.46 / 0.609 \\
&RW-Tensor & 0.075 / 0.242 & 0.025 / 0.249& 0.127 / 0.252  & 0.361 / 0.538 \\
&Gain  & 1.013 / 1.087 & 1.6 / 1.129& 0.913 / 1.107  & 1.274 / 1.132\\
        \midrule
        \parbox[t]{2mm}{\multirow{3}{*}{\rotatebox[origin=c]{90}{SAGE}}}
        &RW-Concat  & 0.021 / 0.192 & 0.03 / 0.176& 0.104 / 0.193  & 0.457 / 0.614\\
&RW-Tensor & 0.023 / 0.201 & 0.04 / 0.177& 0.121 / 0.205  & 0.367 / 0.552 \\
&Gain & 0.913 / 0.955 & 0.75 / 0.994& 0.86 / 0.941 & 1.245 / 1.112  \\
        \bottomrule
    \end{tabular}
    \label{tab:zinc_res}
\end{table*}
Last, we observe that even without message passing layers we are competitive with graph transformers on the Peptides-struct task. This begs the question of to what extent this dataset contains long-range interactions, and more specifically to what extent such interactions can also be captured by strong structural encodings.

\begin{table*}[t!]
    \centering
    \caption{Ablation on Peptides-Struct (MAE). All results are averaged over 3 runs with different seeds.}

    \vspace{2mm}

    \begin{tabular}{rr|cccc}
        \toprule
        \multicolumn{6}{c}{$\mathsf{PeptidesStruct}$ (MAE $\downarrow$)} \\
        \midrule
         \midrule
        \multicolumn{6}{c}{Full message passing} \\
        \midrule
        \parbox[t]{2mm}{\multirow{3}{*}{\rotatebox[origin=c]{90}{$L=1$}}}
& Parameters& 689& 2557& 7169& 162109\\
& RW-Concat& 0.64 / 0.643& 0.64 / 0.643& 0.395 / 0.397& 0.254 / 0.261\\
& RW-Tensor& 0.283 / 0.282& 0.269 / 0.269& 0.255 / 0.26& 0.226 / 0.255\\
        \midrule
        \parbox[t]{2mm}{\multirow{3}{*}{\rotatebox[origin=c]{90}{$L=4$}}}
& Parameters& 1457& 6445& 19457& 477037\\
& RW-Concat& 0.279 / 0.281& 0.264 / 0.267& 0.248 / 0.259& 0.219 / 0.253\\
& RW-Tensor& 0.281 / 0.279& 0.258 / 0.261& 0.247 / 0.257& 0.213 / 0.252\\
        \midrule
        \midrule
                \multicolumn{6}{c}{Sparse message passing $(K=1)$} \\
\midrule
        \parbox[t]{2mm}{\multirow{3}{*}{\rotatebox[origin=c]{90}{$L=1$}}}
& Parameters& 465& 1333& 3201& 57781\\
& RW-Concat& 0.321 / 0.321& 0.398 / 0.399& 0.271 / 0.271& 0.242 / 0.257\\
& RW-Tensor& 0.296 / 0.294& 0.276 / 0.276& 0.27 / 0.27& 0.242 / 0.256\\
        \midrule
        \parbox[t]{2mm}{\multirow{3}{*}{\rotatebox[origin=c]{90}{$L=4$}}}
& Parameters& 561& 1549& 3585& 59725\\
& RW-Concat& 0.294 / 0.291& 0.275 / 0.276& 0.264 / 0.268& 0.229 / 0.252\\
& RW-Tensor& 0.3 / 0.299& 0.272 / 0.272& 0.268 / 0.269& 0.228 / 0.251\\
        
        \midrule
        \midrule
         \multicolumn{6}{c}{No message passing} \\
        \midrule
        \parbox[t]{2mm}{\multirow{3}{*}{\rotatebox[origin=c]{90}{$L=0$}}}
& Parameters& 433& 1261& 3073& 57133\\
& RW-Concat& 0.64 / 0.643& 0.64 / 0.642& 0.527 / 0.528& 0.352 / 0.351\\
& RW-Tensor& 0.302 / 0.299& 0.401 / 0.401& 0.272 / 0.274& 0.248 / 0.26\\
         
        \bottomrule
    \end{tabular}
    \label{tab:ablation}
\end{table*}

\paragraph{ZINC.} The dataset ZINC is a molecular dataset with molecules up to 38 heavy atoms. The task is to predict the penalized logP score, used for training molecular generation models. A subset of 12,000 molecules is used following \citet{dwivedi2020benchmarking}. Results are again reported as train/test pairs and are provided in \autoref{tab:zinc_res}. 

Similarly to the results observed in the Long Range Graph Benchmark, tensor encodings are again mostly on par with concatenation in all scenarios. Moreover, we again observe that sparse MPNN layers are able to perform on par with full weight matrices, suggesting that also in this task there is no need to over-parametrize the message passing layers. Last, we again observe a significant gap in the performance between tensor encodings and concatenation encodings when no MPNN layers are used, again solidifying our intuition that tensor encodings are strictly more expressive encodings for topological information.

\paragraph{Ablation.} 

The ablation study aims to reduce the message passing layers even further by decreasing both the hidden dimension and the number of layers in the message passing phase. The performance is compared again for full message passing, sparse message passing with an entanglement of $K=1$, and no message passing. Moreover, the joint encoder is removed, meaning that the hidden representations in the first layer are simply found by either concatenating or taking the tensor product of the individual encodings. The hidden dimensions considered are $\{16, 36, 64, 328\}$, and compared using one layer $(L=1)$ and four layers $(L=4)$. The results reported are for the Peptides-struct task using a GCN, and are again reported as train/test pairs in \autoref{tab:ablation}.

In the case of full message passing, we observe that we can reduce the number of layers without noticing a significant drop in performance when tensor encodings are used, whereas such a drop is observed in the concatenation case. We also observe that we can reduce the hidden dimension significantly when using tensor encodings without losing much performance, e.g. multiple models with only a few thousand parameters are able to perform almost on par with the $500$K parameter model. This supports the claim that, for this task, structural encoding could be sufficient to leverage the graph structure, and that {\em tensor encodings are significantly more able to capture this information well in sparse and shallow regimes}.

Similarly, when doing sparse message passing, we observe that the effect of reducing the number of layers or hidden dimensions has a much more significant negative effect when the encodings are formed through concatenation than when they are formed through tensor encodings. Finally, again when no message passing is used at all, the tensor encodings are much more robust in the low-parameter regimes.

\paragraph{Is the choice of geometric encoding important?}

The results obtained on ZINC as well as those on Peptides would highlight that when using the full message-passing layer, it seems that there is no significant difference in terms of performance between either of the two, despite the two approaches being geometrically very different. This brings up the question of what the message-passing is actually learning based on the structural information, and to what extent the message-passing layer is actually needed. Our investigation supports the idea that the tensor-product encoding is significantly more resilient to the reduction of the number of parameters in the MPNN, or in fact the removal of the MPNN all together, in contrast to the concatenation approach where instead eliminating the message-passing leads to serious performance degradation. We speculate that one reason for the robustness of the tensor product to the `size' (or presence) of the message-passing as opposed to the concatenation operation, resides in its ability to model more complex and high-dimensional entangled representation which can then be leveraged by an MLP without the usage of the graph.

\section{Discussion}

\paragraph{Conclusion and future research.}

In this work, the effect of the topologically informed node encodings in MPNNs and the role of message passing have been explored. For this, we have proposed a tensor based encoding and compared this against the standard encoding done through concatenation. Moreover, we have looked at the effect of reducing the weights in the MPNN layers or removing the message-passing step all-together, investigating such approaches across different MPNN architectures. 

Our results indicate that when no MPNN layers are included in the model, tensor encodings are able to better learn on graphs in general, and even able to perform on par with models with MPNN layers in some tasks. Moreover, since tensor encodings are almost always at least on par with using standard concatenation, we conclude that tensor encodings form a promising new approach for learning on graphs with structural information. For future research, a natural question is to see if tensor encodings can also improve graph-agnostic architectures such as transformers. 

Moreover, we observe that in the cases where excluding all MPNN layers led to a significant reduction in performance, much of that performance could also be obtained using very sparse weight matrices. This suggests that the role of MPNN layers in these structural tasks is minimal. An interesting avenue to explore relating to this result is to study what MPNNs learn, and whether it is possible to reduce the parameters in MPNN layer.

\bibliography{bibliography}
\bibliographystyle{icml2023}

%%%%%%%%%%%%%%%%%%%%%%%%%%%%%%%%%%%%%%%%%%%%%%%%%%%%%%%%%%%%%%%%%%%%%%%%%%%%%%%
%%%%%%%%%%%%%%%%%%%%%%%%%%%%%%%%%%%%%%%%%%%%%%%%%%%%%%%%%%%%%%%%%%%%%%%%%%%%%%%
% APPENDIX
%%%%%%%%%%%%%%%%%%%%%%%%%%%%%%%%%%%%%%%%%%%%%%%%%%%%%%%%%%%%%%%%%%%%%%%%%%%%%%%
%%%%%%%%%%%%%%%%%%%%%%%%%%%%%%%%%%%%%%%%%%%%%%%%%%%%%%%%%%%%%%%%%%%%%%%%%%%%%%%
\newpage
\appendix
\onecolumn
\section{Detailed results}
\label{app:results_det}

\begin{table*}[h]
    \centering
    \caption{Test results of main experiments reported with standard deviation. All results are averaged over 3 runs with different seeds.}

        \vspace{2mm}

    \begin{tabular}{rcc|cccc}
        \toprule
        \multicolumn{7}{c}{Test performance with standard deviation} \\
        
        \midrule
        &State Type & Layer & full & $K=10$ & $K=1$  & no MP   \\
                \midrule
        \parbox[t]{2mm}{\multirow{6}{*}{\rotatebox[origin=c]{90}{Peptides--Struct}}} & RW-Concat & gcn & $0.255 \pm 0.007 $& $0.255 \pm 0.001$& $0.251 \pm 0.006$& $0.643 \pm 0.000$ \\
& RW-Tensor & gcn & $0.252 \pm 0.001$& $0.254 \pm 0.001$& $0.249 \pm 0.001$& $0.268 \pm 0.002$ \\
& RW-Concat & gin & $0.290 \pm 0.244$& $0.555 \pm 0.000$& $0.252 \pm 0.006$& $0.643 \pm 0.000$ \\
& RW-Tensor & gin & $0.257 \pm 0.005$& $0.309 \pm 0.045$& $0.249 \pm 0.001$& $0.266 \pm 0.001$ \\
& RW-Concat & sage & $0.254 \pm 0.004$&$ 0.256 \pm 0.010$& $0.251 \pm 0.003$& $0.643 \pm 0.000$ \\
& RW-Tensor & sage & $0.253 \pm 0.002$& $0.253 \pm 0.003$& $0.250 \pm 0.001$&$ 0.267 \pm 0.002$ \\
\midrule
                \midrule
               \parbox[t]{2mm}{\multirow{6}{*}{\rotatebox[origin=c]{90}{Peptides-Func}}} & RW-Concat & gcn & $0.613 \pm 0.001$& $0.598 \pm 0.014$&$ 0.606 \pm 0.009$& $0.468 \pm 0.028 $\\
& RW-Tensor & gcn & $0.607 \pm 0.013$& $0.595 \pm 0.010$& $0.593 \pm 0.006$& $0.586 \pm 0.008$ \\
& RW-Concat & gin & $0.577 \pm 0.44$& $0.328 \pm 0.000$& $0.596 \pm 0.016$& $0.490 \pm 0.033$ \\
& RW-Tensor & gin & $0.611 \pm 0.008$& $0.601 \pm 0.003$& $0.606 \pm 0.014$& $0.582 \pm 0.006$ \\
& RW-Concat & sage & $0.607 \pm 0.010$& $0.594 \pm 0.013$& $0.596 \pm 0.009$& $0.501 \pm 0.022$ \\
& RW-Tensor & sage & $0.616 \pm 0.014$& $0.607 \pm 0.005$& $0.601 \pm 0.007$& $0.579 \pm 0.007$ \\
\midrule
\midrule
        \parbox[t]{2mm}{\multirow{6}{*}{\rotatebox[origin=c]{90}{Zinc}}} & RW-Concat & gcn & $0.207 \pm 0.010$& $0.233 \pm 0.008$& $0.24 \pm 0.001$& $0.607 \pm 0.003$ \\
& RW-Tensor & gcn & $0.209 \pm 0.002$& $0.220 \pm 0.001$& $0.219 \pm 0.010$& $0.527 \pm 0.003$ \\
& RW-Concat & gin & $0.263 \pm 0.010$& $0.281 \pm 0.005$& $0.279 \pm 0.008$& $0.609 \pm 0.014$ \\
& RW-Tensor & gin &$ 0.242 \pm 0.008$& $0.249 \pm 0.008$& $0.252 \pm 0.005$& $0.538 \pm 0.008$ \\
& RW-Concat & sage & $0.192 \pm 0.003$& $0.176 \pm 0.014$& $0.193 \pm 0.004$& $0.614 \pm 0.005$ \\
& RW-Tensor & sage & $0.201 \pm 0.003$& $0.177 \pm 0.007$ & $0.205 \pm 0.002$&$ 0.552 \pm 0.006$ \\
        \bottomrule
    \end{tabular}
    \label{tab:std_results}
\end{table*}

%%%%%%%%%%%%%%%%%%%%%%%%%%%%%%%%%%%%%%%%%%%%%%%%%%%%%%%%%%%%%%%%%%%%%%%%%%%%%%%
%%%%%%%%%%%%%%%%%%%%%%%%%%%%%%%%%%%%%%%%%%%%%%%%%%%%%%%%%%%%%%%%%%%%%%%%%%%%%%%

\end{document}